\documentclass[letterpaper]{article} 
\usepackage{aaai25}  
\usepackage{times}  
\usepackage{helvet}  
\usepackage{courier}  
\usepackage[hyphens]{url}  
\usepackage{graphicx} 
\usepackage{natbib}  
\usepackage{caption} 
\usepackage{algorithm}
\usepackage{algpseudocode}

\usepackage{booktabs}
\usepackage{makecell}
\usepackage{multirow}
\usepackage{amsmath}
\usepackage{amsfonts}
\usepackage{bbding}
\usepackage{newtxmath}
\usepackage{pifont}
\usepackage{float} 

\usepackage{newfloat}
\usepackage{listings}
\DeclareCaptionStyle{ruled}{labelfont=normalfont,labelsep=colon,strut=off} 
\floatstyle{ruled}
\newfloat{listing}{tb}{lst}{}
\floatname{listing}{Listing}
\title{CREST: An Efficient Conjointly-trained Spike-driven Framework for Event-based Object Detection Exploiting Spatiotemporal Dynamics}
\author{
    Ruixin Mao\equalcontrib,
    Aoyu Shen\equalcontrib,
    Lin Tang ,
    Jun Zhou\thanks{Corresponding author: Jun Zhou}
}
\affiliations{
    University of Electronic Science and Technology of China,  Chengdu 611731, China\\

    \{mao.rx, shen\_aoyu, tangl\}@std.uestc.edu.cn, zhouj@uestc.edu.cn
}

\begin{document}

\maketitle

\begin{abstract}

Event-based cameras feature high temporal resolution, wide dynamic range, and low power consumption, which are ideal for high-speed and low-light object detection.  Spiking neural networks (SNNs) are promising for event-based object recognition and detection due to their spiking nature but lack efficient training methods, leading to gradient vanishing and high computational complexity, especially in deep SNNs. Additionally, existing SNN frameworks often fail to effectively handle multi-scale spatiotemporal features, leading to increased data redundancy and reduced accuracy. To address these issues, we propose CREST, a novel conjointly-trained spike-driven framework to exploit spatiotemporal dynamics in event-based object detection. We introduce the conjoint learning rule to accelerate SNN learning and alleviate gradient vanishing. It also supports dual operation modes for efficient and flexible implementation on different hardware types. Additionally, CREST features a fully spike-driven framework with a multi-scale spatiotemporal event integrator (MESTOR) and a spatiotemporal-IoU (ST-IoU) loss. Our approach achieves superior object recognition \& detection performance and up to 100$\times$ energy efficiency compared with state-of-the-art SNN algorithms on three datasets, providing an efficient solution for event-based object detection algorithms suitable for SNN hardware implementation. 
\end{abstract}

\begin{links}
\link{Code}{https://github.com/shen-aoyu/CREST/}
\end{links}

\section{Introduction}
Frame-based cameras excel in capturing photometric features but suffer from motion blur with fast-moving objects and are sensitive to extreme lighting conditions. Continuous full-frame processing also consumes high energy due to redundant background information. Conversely, event-based cameras capture changes in light intensity asynchronously at each pixel \cite{tobi2008dvs, tobi2020event}. This allows them to operate at much higher frequencies and under varied illumination conditions, ideal for high-speed and real-time object detection.

Deep ANN frameworks for event-based object detection encode event streams into dense, image-like representations, leveraging conventional frame-based computer vision techniques \cite{peng2023dvsann, zubic2024dvsann}. Modern GPUs and TPUs also accelerate ANN training and inference \cite{nickolls2008gpu, jouppi2017tpu}. However, ANNs struggle with effectively handling the temporal features inherent in event-based data. This arises from losing crucial temporal dynamics and correlations when encoding events into discrete frames. Additionally, they consume high energy due to numerous MAC operations, which is problematic for energy-sensitive edge applications.

SNNs mimic the spiking feature of the biological neurons which are theoretically well-suited for processing event-based data \cite{maass1997snn,roy2019snn}. The timing and frequency of spikes convey diverse information, granting SNNs strong spatiotemporal characteristics. Moreover, SNNs perform accumulations (ACs) only when sparse spikes occur, making them inherently energy-efficient. However, SNNs often struggle with deep network structures due to inefficient training methods \cite{zhang2024shallowsnn, nagaraj2023dotie}. Non-differentiable spikes require surrogate gradients for backpropagation (BP), potentially causing gradient vanishing in deep SNNs \cite{wu2018stbp, neftci2019bpsnn}. The propagation of spatiotemporal gradients also greatly increases computation complexity. Some works convert well-trained ANN models to the same structured SNN. However, they require a large encoding time-window to approximate the ANN performance, causing a much higher spiking rate and energy consumption \cite{kim2020spikingyolo, li2022converObjectdetection}. More complex spiking neurons and layers with trainable parameters are proposed to lower the spiking rate, which add computation complexity for training \& inference and are difficult to implement on existing SNN hardware \cite{fang2021plif, cordone2022object}. 

Furthermore, a unified SNN framework is needed to handle the multi-scale spatiotemporal features of event-based object detection. Existing approaches often fail to effectively encode the spatiotemporal characteristics of event data, leading to data redundancy and reduced accuracy \cite{cordone2022object, bodden2024spikingcenternetdistillationboostedspiking}. Moreover, the Intersection over Union (IoU) loss \cite{yu2016iou} employed in SNNs are generally adapted from ANNs, neglecting the unique spatiotemporal nature of event-based spike trains\cite{su2023deep, fan2024sfod}. This may lead to inaccurate regression and thus decrease the detection accuracy. 

To this end, we propose \textit{CREST}, an efficient conjointly-trained spike-driven framework exploiting spatiotemporal dynamics for event-based object detection. Firstly, we propose a simple yet effective conjoint learning rule with dual operation modes for efficient and flexible training implementation on different hardware types. For the backward process, we design a surrogate neural network with discrete-level activation values (DL-Net) to mimic the values represented by different spike train patterns.  This replaces the original spatiotemporal gradient calculations, which reduces the computation complexity, alleviates the gradient vanishing problem, and speeds up the learning process compared with traditional SNN BP-like or conversion-based training. 
 
Additionally, we propose a fully spike-driven framework for event-based object detection which includes a multi-scale spatiotemporal event integrator (MESTOR), a spatiotemporal-IoU (ST-IoU) loss, and few-spikes neuron (FSN) \cite{stockl2021optimized} based SNN model. MESTOR not only aggregates event data across multiple scales but also extracts the spatiotemporal continuous events. This keeps the key spatiotemporal feature and reduces redundant background and noise events meantime. ST-IoU loss comprises the proposed spiking density-based IoU (Spiking-IoU)  to exploit spatiotemporal continuity and the Complete-IoU (CIoU) \cite{zheng2021ciou} for coordinate loss. The FSN-SNN adopts an efficient spike encoding scheme and is supported by the recently proposed high-performance SNN hardware STELLAR \cite{mao2024stellar}. 

Our main contributions can be summarized as follows:
\begin{itemize}
    \item \textbf{\textit{CREST}, a spike-driven conjointly-trained framework exploiting the spatiotemporal dynamics} to enhance the efficiency of event-based object detection.
    \item \textbf{A novel conjoint learning rule} which introduces a surrogate neural network with discrete-level activation values to accelerate the learning process and alleviate the gradient vanishing issues in deep SNNs. Dual operation modes add flexibility to its hardware implementation.
    \item \textbf{A fully spike-driven framework} which incorporates the MESTOR, ST-IoU loss, and FSN-SNN model to handle multi-scale spatiotemporal features of the event-based object recognition and  detection.
    \item Compared with the  state-of-the-art SNN algorithms, our work achieves superior recognition \& detection performance  and up to 100$\times$ energy efficiency on 3 datasets.
\end{itemize}

\section{Background and Motivation} \label{background}

\subsubsection{Inefficiencies of LIF-like spiking neuron models.} Leaky Integrate-and-Fire (LIF) neuron and its variants (details are shown in\textbf{ Supplementary Material A}) are most commonly used in SNN algorithms and hardware implementations due to their trade-off between low computational complexity and biological interpretability \cite{abbott1999lif, gerstner2002IF,fang2021plif}. However, LIF neurons only support temporal or frequency encoding, which requires a long time window and many spikes to achieve high accuracy. Some introduce trainable parameters and normalization functions  (e.g. complex exponent and division operations) to lower the spiking rate. However, the existing SNN hardware designs mainly adopt IF neurons for high energy efficiency which cannot support these complex operations.

\begin{figure*}[tb]
    \centering    
    \includegraphics[width=0.85\textwidth]{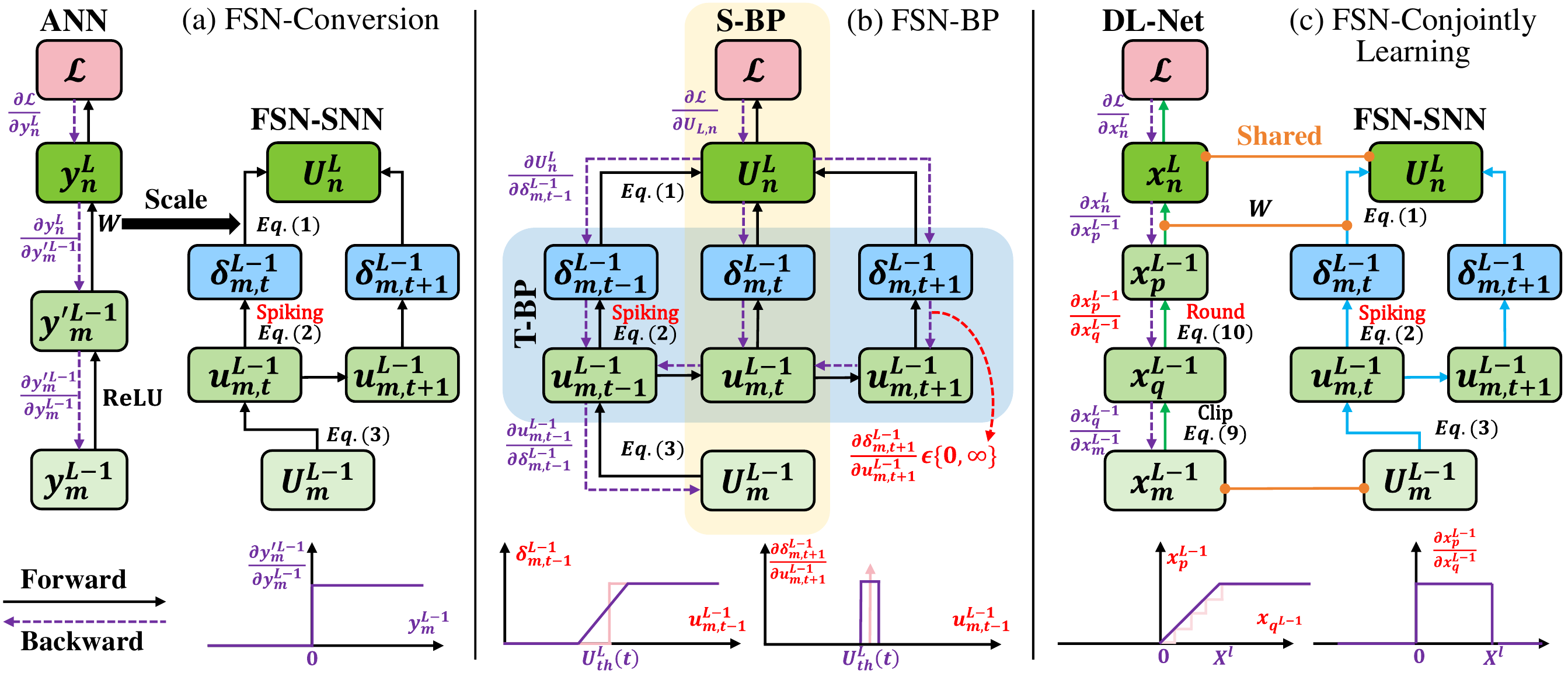}
    \caption{Comparison of (a) FSN-conversion, (b) FSN-BP, and (c) FSN-conjointly learning. The last layer uses the accumulated membrane potentials to make decisions. S-BP/T-BP denotes spatial-BP/temporal-BP. The light red/purple line represents the original/approximated gradient.} 
    \label{fig.learn}
\end{figure*}

\subsubsection{Efficient Few-Spikes neuron (FSN).} 
To improve encoding efficiency, the FSN and the FSN-based ANN-to-SNN conversion learning (FSN-conversion) method are proposed \cite{stockl2021optimized}. The spike trains encoded by FSNs carry both temporal and frequency information, which convey more information within a shorter time window than
LIF-like neurons. Additionally, FSN-based SNN hardware demonstrates much higher energy efficiency and speedup compared to conventional SNN hardware \cite{mao2024stellar}. Therefore, we adopt FSNs as foundational units in this work. 

The FSN works as follows. Firstly, the FSN $n$ in layer $l$ integrates the input spike $\delta_m^{l-1}$ to the membrane potential $U_n^l$ (Eq. \ref{eq.integration}). $N^{l-1}$, $K^{l-1}$, $d^{l-1}(t)$, $w_{mn}^l$ represent the number of input neurons, time-window length, spike coefficient at time $t$, and connection weights respectively. If $u_n^l(t)$ exceeds the threshold $U_{th}^{l}(t)$, a spike is fired (Eq. \ref{eq.spike}), and $u_n^l(t)$ decays by $U_{th}^{l}(t)$ (Eq. \ref{eq.membraneDecay}). The neuron dynamics parameters $U_{th}^{l}(t)$ and $d^l(t)$ can be used to model various activation functions.

\begin{align}
 \label{eq.integration}U_{n}^l &=\sum_{m=1}^{N^{l-1}}\sum_{t=1}^{K^{l-1}}d^{l-1}(t)\delta_m^{l-1}(t)w_{mn}^l\\
\label{eq.spike}\delta_n^l(t) &= \begin{cases}1, & u_n^l(t) \ge U_{th}^{l}(t) \\ 0, & \text {otherwise }\end{cases} \\
\label{eq.membraneDecay}u_n^l(t+1)&=U_n^l-\sum_tU_{th}^{l}(t)\delta_n^l(t)   
\end{align}

\begin{algorithm}[t]
\small
\caption{FS-Neuron based Conjoint Learning}
\label{alg:algorithm}
\textbf{Input}: model weight $w_{mn}^l$ from layer $l-1$ to $l$, input membrane potential $U_m^{l-1}/x_m^{l-1}$, threshold $\alpha^l$ and and time-window $K^l$ of layer $l$\\
\textbf{Output}: Updated model weight $\textbf{W}$ 
\begin{algorithmic}[1] 
\State \textbf{Forward Pass:}
\If {$SNN-forward$} \texttt{//In SNN Hardware}
\For{$t=1$ to $K^l$}
\State $\delta_m^{l-1}(t)$ $\leftarrow$ $Spiking(u_m^{l-1}(t),\alpha^{l-1},K^{l-1})$ \Comment{Eq.(\ref{eq.spike})}
\EndFor
\State $x_n^l=U_n^{l}$ $\leftarrow$ $ Integ.(\delta_m^{l-1}(t),\alpha^{l-1},K^{l-1},w_{mn}^l)$ \!\!\!\!\Comment{Eq.(\ref{eq.integration})}
\Else    \texttt{   //In GPU/TPU}
\State $x_q^{l-1} \leftarrow Clip(x_m^{l-1},\alpha^{l-1},K^{l-1})$ \Comment{Eq.(\ref{eq.dln_clip})}
\State $x_p^{l-1} \leftarrow Round(x_q^{l-1},\alpha^{l-1},K^{l-1})$ \Comment{Eq.(\ref{eq.dln_quantilize})}
\State $x_n^{l} \leftarrow Convolution(x_p^{l-1},w_{mn}^l)$ \Comment{Eq.(\ref{eq.dln_integrate})}
\EndIf
\State \textbf{Backward Pass:}
\State \texttt{//Surrogate gradient BP in DL-Net}
\State $\frac{\partial x_{n}^{l}}{\partial x_{m}^{l-1}}=\frac{\partial x_{n}^{l}}{\partial x_{p}^{l-1}} \frac{\partial x_{p}^{l-1}}{\partial x_{q}^{l-1}}\approx\frac{\partial x_{n}^{l}}{\partial x_{p}^{l-1}}$
\State \textbf{return} Updated model weight $\textbf{W}$ 
\end{algorithmic}
\end{algorithm}

\subsubsection{Inefficiencies of the conventional SNN training methods.} \underline{FSN-conversion (Fig. \ref{fig.learn}(a)):} It uses neuron dynamics parameters to emulate the ANN activation functions, allowing the weights from a trained ANN to be directly applied to the same-structured SNN. Despite its better performance than LIF-based SNNs, it still limits the accuracy with the short time-window due to the difference between ANN and FSN functions. Increasing the time-window improves the approximation between FSN and ANN functions thus reducing the conversion accuracy loss. But this also increases spiking rate and energy consumption as demonstrated in \cite{mao2024stellar}. \underline{FSN based BP (FSN-BP) learning (Fig. \ref{fig.learn}(b)):} Similar to the traditional SNN BP-like learning, FSN-BP can be achieved with a surrogate gradient for non-differentiable spike. The spatiotemporal backpropagation is denoted as Eq. \ref{eq.fsbp}, with $\epsilon_m^{L-1}$ as error from loss $\mathcal{L}$ to layer $L-1$. The surrogate gradient is approximated as Eq. \ref{eq.fsspikegrad}, where $\text{rect}$ equals 1 when the input is within $[-0.5,0.5]$ and 0 otherwise. This approach achieves higher accuracy with a shorter time-window and fewer spikes than FSN-Conversion. However, it has high computation complexity due to the spatiotemporal gradient calculation. Further, deep networks exacerbate the issue of handling and storing sparse spike timing information. Gradients become very small and even diminish during backpropagation, pronouncing the gradient vanishing problem. 

\begin{equation}\label{eq.fsbp}
\epsilon^{L-1}_m = \frac{\partial\mathcal{L}}{\partial U_m^{L-1}}=
\sum_{n=1}^{n^L}\sum_{t=1}^{K^L}  w^{L}_{mn}\frac{\partial \delta^{L-1}_m(t)}{\partial u^{L-1}_m(t)} d^L(t) \epsilon^L_n
\end{equation}

\begin{equation}\label{eq.fsspikegrad}
\frac{\partial \delta^{L-1}_{m}(t)}{\partial u^{L-1}_m(t)} = \frac{1}{2\beta} \cdot \text{rect}\left(\frac{u^{L-1}_m(t) - U^{L-1}_{th}(t)}{2\beta}\right)
\end{equation}

\section{Method}
We propose \textit{CREST}, a spike-driven conjointly-trained framework exploiting the spatiotemporal dynamics. We establish a dual-model conjoint learning rule to simplify the spatiotemporal BP which support efficient implementation on different hardware types. Based on this, a fully spike-driven framework is developed to handle multi-scale spatiotemporal features of event-based object detection. 

\subsection{FS-Neuron based Conjoint Learning Rule}
We first demonstrate how to use FSN to emulate ReLU to give an insight into the FSN computation mechanism. Then, a surrogate DL-Net is proposed to mimic the continuous values represented by discrete spike train patterns. Further, we introduce the spatial surrogate gradient backpropagation to DL-Net. Finally, we illustrate the operation flow of the conjoint learning rule.

\subsubsection{Similarity between FSN and ReLU Functions.} ANN-based object detection mostly adopts ReLU and leaky-ReLU \cite{glorot2011relu, maas2013leakrelu}. We implement ReLU in this paper and the detailed reasons are illustrated in \textbf{Supplementary Material B}. Based on Eq. \ref{eq.membraneDecay}, rewrite the membrane potential $U_{m}^{l-1}$ as Eq. \ref{eq.membrane_nout}. Meanwhile, denote the unweighted integration of output spikes from FSN $m$ as Eq. \ref{eq.integration_nout} based on Eq. \ref{eq.integration}.

{\fontsize{9.5pt}{6.5pt}\selectfont 
\setlength{\abovedisplayskip}{0pt} 
\setlength{\belowdisplayskip}{0pt} 
\setlength{\abovedisplayshortskip}{1pt} 
\setlength{\belowdisplayshortskip}{1pt} 
\begin{equation}\label{eq.membrane_nout}
\begin{aligned}
U_{m}^{l-1}=\sum_{t=1}^{K^{l-1}}U_{th}^{l-1}(t)\delta_m^{l-1}(t)+u_m^{l-1}(K^{l-1})
\end{aligned}
\end{equation}}
{\setlength{\abovedisplayskip}{0pt}%
\setlength{\belowdisplayskip}{0pt}%
\setlength{\abovedisplayshortskip}{0pt}%
\setlength{\belowdisplayshortskip}{1pt}%
\begin{equation}\label{eq.integration_nout}
U_{p}^{l-1}=\sum_{t=1}^{K^{l-1}}d^{l-1}(t)\delta_m^{l-1}(t)
\end{equation}
}

\begin{figure*}[tb]
    \centering
    \includegraphics[width=0.9\textwidth]{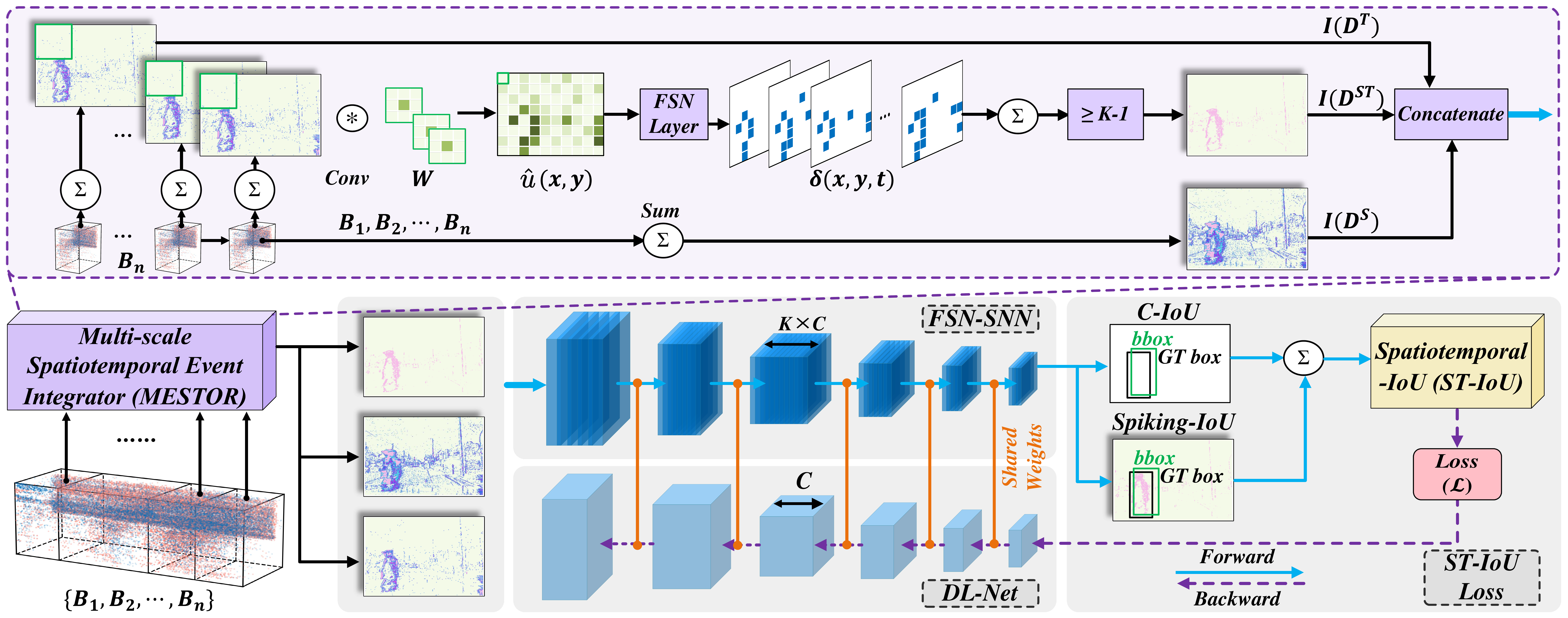}
    \caption{The overview architecture of \textit{CREST}.}
    \label{fig.arch}
\end{figure*}

With the help of FSN parameters, the membrane potential is encoded into a spike train and then decoded into a continuous value. In this process, $U_{m}^{l-1}$ and $U_{p}^{l-1}$ emulate the input and output of ReLU, and the MAC in ANNs are replaced by the ACs of the corresponding weight at spiking time steps in SNNs. To preserve non-linearity while controlling the output range, ReLU is bounded by $\alpha$. This is formulated as three cases: \ding{182} \!when $U_{m}^{l-1}\!\leq\!0$, $U_p^{l-1}\!=\!0$; \ding{183} \!when $0\!<\!U_{m}^{l-1}\!\leq \!\alpha$, $U_p^{l-1}\!=\!U_{m}^{l-1}$; and \ding{184} \!when $U_{m}^{l-1}\!>\!\alpha$, $U_p^{l-1}\!\!=\!\alpha$. Case \ding{183} is optimized by minimizing the error with proper FSN parameters under the distribution of $U_{m}^{l-1}$ (Eq. \ref{eq.opti3}), with $K^{l-1}$ pre-set by the requirements of accuracy and firing rate.

\begin{equation}\label{eq.opti3}
\textbf{min}\; \varepsilon(U_{th}^{l-1}(t),d^{l-1}(t);K^{l-1})=\mathbb{E}_{u_m(t)}|U_m^{l-1}-U_p^{l-1}|
\end{equation}

A possible setting is $U_{th}^{l-1}(t)\!\!=\!d^{l-1}(t)\!\!=\!\alpha^{l-1}2^{-t}$\!. For case \ding{182}, when $U_{m}^{l-1}\!\le\!0$, FSN $m$ \!generates no spikes ($U_p^{l-1}\!\!=\!0$). For case \ding{183}, when $U_{m}^{l-1}$ is an integer multiple of $\alpha^{l-1}2^{-K^{l-1}}$, $U_p^{l-1}\!=\!U_{m}^{l-1}$. Otherwise, $U_m^{l-1}$ rounds down to the nearest level with a tolerance  $\alpha^{l-1}2^{-K^{l-1}}$.  For case \ding{184}, $U_m^{l-1}(t)\!\!\ge\!\alpha^{l-1}2^{-t}$, the spike train is encoded as $K$-bit 1 with $U_{p}^{l-1}\!=\!\alpha^{l-1}(1\!-\!2^{-K^{l-1}})$. The error between $U_p^{l-1}$ and $U_m^{l-1}$ decreases when $K^{l-1}$ increases.  Noted that both the earlier spikes and the higher spiking frequency represent larger values (more significant information), as reflected by larger coefficients ($d(t)$) and accumulated values, which improves encoding efficiency.

\vspace{-0.25em}

\subsubsection{Surrogate DL-Net.} FSN-BP requires complex and iterative spatiotemporal gradient BP with high computation complexity, which can exacerbate gradient vanishing in deep SNNs. Inspired by the coarse-to-fire-processing strategy of FSN-ReLU, we design a surrogate neural network with discrete-level activation values (DL-Net) to approximate spike train patterns in Fig. \ref{fig.learn}(c). 

To emulate the boundary condition of the 3 cases, $x_m^{l-1}$ is clipped to $x_q^{l-1}$ (Eq. \ref{eq.dln_clip}) with $X^{l-1}\!=\!\alpha^{l-1}(1-2^{-\!K^{l-1}}\!)$. Further, $x_q^{l-1}$ is discretized to mimic the FSN spiking process with $X^{l-1}_\text{min}\!\!=\alpha^{l-1}2^{-K^{l-1}}$ (Eq. \ref{eq.dln_quantilize}).  Finally, the temporal iterative ACs in Eq. \ref{eq.integration} are replaced by MACs (Eq. \ref{eq.dln_integrate}), where $x_n^l$ emulates $U_n^l$. The connecting weights $w_{pn}^l$ in DL-Net are shared with those in the FSN-SNN.  

{\setlength{\abovedisplayskip}{1pt}%
\setlength{\belowdisplayskip}{1pt}%
\setlength{\abovedisplayshortskip}{1pt}%
\setlength{\belowdisplayshortskip}{1pt}%
{\begin{equation}\label{eq.dln_clip}
x_q^{l-1}=\text{clip}(x^{l-1}_m)=\begin{cases} 
0, &  x^{l-1}_m < 0 \\
x^{l-1}_m, &  0\leq x^{l-1}_m\leq X^{l-1} \\
X^{l-1}, & \text{otherwise}
\end{cases}
\end{equation}}}
{\setlength{\abovedisplayskip}{1pt}%
\setlength{\belowdisplayskip}{1pt}%
\setlength{\abovedisplayshortskip}{1pt}%
\setlength{\belowdisplayshortskip}{1pt}%
{\begin{equation}\label{eq.dln_quantilize}
x_p^{l-1}= \text{round}(\frac{x_q^{l-1}}{X^{l-1}_\text{min}})\times X^{l-1}_\text{min}
\end{equation}}}
{\setlength{\abovedisplayskip}{1pt}%
\setlength{\belowdisplayskip}{1pt}%
\setlength{\abovedisplayshortskip}{1pt}%
\setlength{\belowdisplayshortskip}{1pt}%
{\begin{equation}\label{eq.dln_integrate}
x^l_n=\sum_{p=1}^{N^{l-1}}x_p^{l-1}w_{pn}^l
\end{equation}}}

\subsubsection{Spatial surrogate gradient backpropagation in DL-Net}
The round function in Eq. \ref{eq.dln_quantilize} causes abrupt changes, leading to a discontinuous derivative. We replace the round function with a smooth identity function during backpropagation as shown in the bottom of Fig.\ref{fig.learn} (c). Therefore, the iterative surrogate gradient calculation for spatiotemporal spiking in Eq. \ref{eq.fsbp} is replaced by the surrogate gradient  (Eq. \ref{eq.dln_spikegrad}) for spatial discretization, which greatly reduces the computation complexity and improves training efficiency. Additionally, the sparse spikes are aggregated into a continuous value, which avoids small and unstable gradients, greatly alleviating the gradient vanishing problem. The detailed process of BP in DL-Net is illustrated in  \textbf{Supplementary Material C}.

\begin{equation} \label{eq.dln_spikegrad}
\frac{\partial x_p^l}{\partial x_q^{l}} \approx 1
\end{equation}

\subsubsection{Dual training modes for flexible hardware implementation (Algorithm 1).} \underline{Case A: GPU/TPU implementation.} GPUs/TPUs primarily focus on accelerating parallel matrix operations, prioritizing speed over energy efficiency. Thus, DL-Net is used for forward \& backward passes in training (green \& purple line in Fig.\ref{fig.learn} (c))).  \underline{Case B: SNN Hardware.}  FSN-based SNN hardware is specifically designed to accelerate FSN computation with ultra-low energy consumption. Therefore, FS-SNN/DL-Net can be used for the forward/backward pass (blue/purple line), which share membrane potentials and weights (oringe line).

\subsection{Overview of Spike-driven Framework for Event-based Object Detection}
Fig. \ref{fig.arch} shows the overall architecture of the spike-driven framework for event-based object detection which includes MESTOR, ST-IoU loss, and FSN-SNN model. MESTOR integrates the input feature from multi-scales. The integrated features are then sent to the FSN-based SNN structure. The FSN model is chosen based on its high encoding efficiency, efficient hardware implementation \cite{mao2024stellar}, and the proposed efficient conjointly training method mentioned above. Additionally, the ST-IoU loss is proposed to improve detection accuracy for the spatiotemporal event-based data.

\begin{table*}[t]
\centering
\label{tab:example}
\setlength\tabcolsep{3.8pt} 
\fontsize{9pt}{8pt}\selectfont
\begin{tabular}{ccccccccc}
\toprule

Method & Representation & Net & Params & Acc & K & fr/spar & AC/MAC & Energy(mJ)\\
\midrule

YOLE \cite{cannici2019asynchronous}  & VoxelGrid & ANN & - & 0.927 & - & 1 & 0.16G & 0.75   \\
EvS-S \cite{li2021graph}& Graph & GNN & - & 0.931 & - & - &  -  & - \\
Asynet \cite{messikommer2020event}& VoxelGrid & ANN & - & 0.944 & - & \textbf{0.067} &  0.16G  & 0.05  \\

\midrule
HybridSNN  \cite{kugele2021hybrid}& HIST & SNN & - & 0.770 & - & - & - & -   \\
Gabor-SNN \cite{sironi2018hats}& HAT & SNN & - & 0.789 & - & - & - & -   \\
Squeeze-1.1 \cite{cordone2022object}& VoxelCube & SNN & \textbf{0.72M} & 0.846 & 5 & 0.251 & 0.02G & 0.36   \\
Mobile-64 \cite{cordone2022object}& VoxelCube & SNN & 18.81M & 0.917 & 5 & 0.171 & 4.20G & 83.34   \\
Dense121-24 \cite{cordone2022object}& VoxelCube & SNN & 3.93M & 0.904 & 5 & 0.336 & 2.25G & 37.76   \\
VGG-11 \cite{cordone2022object}& VoxelCube & SNN & 9.23M & 0.924 & 5 & 0.120 & 0.61G & 12.68   \\

DenseNet121-16 \cite{fan2024sfod}& VoxelCube & SNN & 1.76M & 0.937 & 5 & 0.147 & 0.06G* & 1.22  \\

\midrule

\textit{CREST} (CSPdarknet-tiny) & MESTOR & SNN & 3.41M & 0.949 & 5 & 0.165 & 0.05G & 0.04  \\
\textit{CREST} (DenseNet121-16) & MESTOR & SNN & 1.95M &  \textbf{0.952} & 5 & \textbf{0.146} & 0.07G & 0.04 \\
\textit{CREST} (ShuffleNetV2) & MESTOR & SNN & \textbf{1.01M} & 0.940 & 5 & 0.176 & \textbf{0.01G} & \textbf{0.01}  \\
\bottomrule
\multicolumn{9}{l}{\makecell[l]{{
\small{$^*$ denotes self-implementation result; Acc denotes Accuracy.}
 }}}
\end{tabular}
\caption{Comparison between \textit{CREST} and the state-of-the-art methods on NCARs dataset.} \label{tab.ncar}
\end{table*}

\subsection{Multi-scale Spatiotemporal Event Integrator}
For a continuous event-based period $\mathcal{I}$,  the raw stream $E\!\!=\!\!{e_i(x_i,y_i,t_i,p_i)}_{i\in M}$\! with total $M$ bipolar events $p_i$ at pixel point $(x_i,y_i)$ exhibits high sparsity and temporal redundancy.  These characteristics impede feature learning and increase computational overhead. Moreover, the raw stream contains many redundant background and noise events, which can disrupt the learning process. Existing methods mostly split $\mathcal{I}(E)$ into equal bins and integrate the events in each bin to obtain a spatiotemporal dense representation (Eq. \ref{eq.STrep}) \cite{su2023deep,fan2024sfod}. The bin length affects the spatiotemporal feature preservation, leading to a trade-off between both scales. We aim to integrate the input feature from multi-scales (spatiotemporal scale $\mathcal{ST}$, spatial scale $\mathcal{S}$ and temporal scale $\mathcal{T}$) for better representation.

\begin{align}
    \label{eq.STrep}\mathcal{I}(E) &\Longrightarrow \mathcal{I}(D^{\mathcal{ST}})
\end{align}

\subsubsection{Spatiotemporal continuous $\mathcal{I}(D^{\mathcal{ST}})$.} We observe that the events generated by the same object exhibit both time and space continuity, whereas background and noise events are less time-continuous. Therefore, we leverage the spatiotemporal properties of the spiking neuron to extract and cluster the spatiotemporal continuous events and reduce redundant background and noise events. Like \cite{nagaraj2023dotie}, we first divide $\mathcal{I}(E)$ into $N$ equal-length ($\Delta t$) time bins $\mathcal{B}(x,y,n)$ without polarity. 
\setlength{\abovedisplayskip}{1pt}%
\begin{align}
\label{eq.Timewindow} 
\mathcal{B}(x,y,n) &= \sum_{i=1}^{(n-1) \Delta t < e(t_i) < n \Delta t} e_i(x_i, y_i, t_i)
\end{align}
\setlength{\belowdisplayskip}{0pt}%

An FSN-based convolution layer is constructed which applies fixed-value filters of size $3\times3$ to time bin $\mathbf{\mathcal{B}}_t$. In detail, the convolution results of each time bin and each filter are accumulated to the membrane potential $\widehat{u}$ in each time step (Eq. \ref{eq.mestormembrane}). Then, based on Eq.\ref{eq.spike} and \ref{eq.membraneDecay}, each neuron generates a spike train (Eq. \ref{eq.mestorspike}). With a given time window $K$, if the number of spikes in one spike train is greater than $K-1$, this neuron (pixel) is considered space-time continuous and we set the value of this pixel as 1 (we retain this pixel). Otherwise, we abandon this pixel (Eq. \ref{eq.memstorevent}).
\setlength{\abovedisplayskip}{1pt}%
\begin{align}
\label{eq.mestormembrane} 
\widehat{u}(x,y,t) &= \mathbf{\mathcal{B}}_t * \mathbf{W} + u(x,y,t) \\
\label{eq.mestorspike} 
\{\delta(x, y, t),  u(x,y&,t+1)\} = \text{FSN}(\widehat{u}(x,y,t)) \\
\label{eq.memstorevent} 
\mathcal{I}(D^{\mathcal{ST}})(x, y) &= \begin{cases} 
1, & \sum_{t=1}^{K} \delta(x, y, t) \geq K-1 \\
0, & \text{otherwise} 
\end{cases}
\end{align}
\setlength{\belowdisplayskip}{0pt}%
\subsubsection{Spatial scale $\mathcal{I}(D^{\mathcal{S}})$ and temporal scale $\mathcal{I}(D^{\mathcal{T}})$.} Combined with Eq. \ref{eq.Timewindow}, we observe that shorter time bins capture finer temporal details but increase data sparsity, while longer intervals enhance spatial feature density but may smooth out crucial temporal features. So we use short time bins to keep temporal features (Eq. \ref{eq.mestortem}) and use the accumulation of short time bins to keep spatial features (Eq. \ref{eq.memstorsp}). 
\setlength{\abovedisplayskip}{1pt}%
\begin{align}
    \label{eq.mestortem} \mathcal{I}(D^{\mathcal{T}})(x,y) & = \mathcal{B}(x,y,N) \\ 
    \label{eq.memstorsp} \mathcal{I}(D^{\mathcal{S}})(x,y) &=\sum_{n=1}^{N} \mathcal{B}(x,y,n) 
\end{align}
\setlength{\belowdisplayskip}{0pt}%

Finally, the features extracted from multi-scales are integrated as a 3-channel input and fed into the following network as Eq. (\ref{eq.decoupleSTrep}).
\setlength{\abovedisplayskip}{1pt}%
\begin{align}
    \label{eq.decoupleSTrep}\mathcal{I}(E) &\Longrightarrow  \mathcal{I}(D^{\mathcal{ST}}) \; \& \; \mathcal{I}(D^{\mathcal{S}}) \; \& \; \mathcal{I}(D^{\mathcal{T}}) 
\end{align}
\setlength{\belowdisplayskip}{0pt}%

\begin{table*}[t]
\centering
\fontsize{9pt}{8pt}\selectfont
\setlength\tabcolsep{1.2pt} 
\begin{tabular}{ccccccccccc}
\toprule
Method & Rep. & Net & Head & Params &  mAP$_{50:95}$ & mAP$_{50}$ & K & fr/sp & AC/MAC & Energy\\
\midrule
Asynet \cite{messikommer2020event}& VoxelGrid & ANN & YOLO & 11.4M & 0.129 & - &  -  & 0.100 & 1.05G & 0.48\\
S-Center \cite{bodden2024spikingcenternetdistillationboostedspiking}& HIST & ANN & CenterNet & 12.97M & 0.278 & - &  -  & 1 & 6.13G & 28.21\\
EGO-12 \cite{zubic2023chaos}& EGO-12 & ANN & YOLOv6 & 140M* & \textbf{0.504} & - &  -  & 1 & 84.34G* & 387.96\\
\midrule
VC-Dense \cite{cordone2022object}& VoxelCube & SNN & SSD & 8.2M & 0.189 & - &  5  & 0.372 & \textbf{2.33G} & 37.55\\
VC-Mobile \cite{cordone2022object}& VoxelCube & SNN & SSD & 24.26M & 0.147 & - &  5  & 0.294 & 4.34G & 76.22\\
LT-SNN \cite{hasssan2023ltsnn}& HIST & SNN & YOLOv2 & 86.82M* & 0.298 & - &  -  & - & 19.53G* & -\\
S-Center \cite{bodden2024spikingcenternetdistillationboostedspiking}& HIST & SNN & CenterNet & 12.97M & 0.229 & - &  5  & 0.174 & 6.38G & 126.21\\
EMS-10 \cite{su2023deep}& HIST & SNN & YOLOv3 & 6.20M & 0.267 & 0.547 &  5  & 0.211 & 5.90G* & 112.83\\
EMS-18 \cite{su2023deep}& HIST & SNN & YOLOv3 & 9.34M & 0.286 & 0.565 &  5  & 0.201 & 9.70G* & 187.02\\

EMS-34 \cite{su2023deep}& HIST & SNN & YOLOv3 & 14.40M & 0.310 & 0.590 &  5  & 0.178 & 32.99G* & 650.13\\
TR-YOLO \cite{yuan2024trainable}& HIST & SNN & YOLOv3 & 8.7M & - & 0.451 &  3  & - & - & -\\
SFOD \cite{fan2024sfod}& VoxelCube & SNN & SSD & 11.9M & 0.321 & 0.593 &  5  & 0.240 & 6.72G* & 124.73\\
\midrule
\textit{CREST} (CSPdarknet-tiny)  & \multirow{3}{*}{MESTOR} & SNN & YOLOv4 & 7.61M & \textbf{0.360} & \textbf{0.632} & 5 & 0.167 & 8.39G & 6.31\\
\textit{CREST} (DenseNet121-16) &  ~ & SNN & YOLOv4 & 3.01M & 0.339 & 0.615  & 5 & \textbf{0.095} & 8.15G& 3.48\\
\textit{CREST} (ShuffleNetV2) & ~ & SNN & YOLOv4 &  \textbf{1.79M} & 0.305 & 0.568  & 5 & 0.208 & \textbf{2.42G}& \textbf{2.27}\\

\bottomrule
\multicolumn{11}{l}{\makecell[l]{{
\small{$^*$ denotes self-implementation results; Rep. denotes Representation; energy unit is mJ.}
 }}}
\end{tabular}
\caption{Comparison between \textit{CREST} and the state-of-the-art methods on Gen1 dataset.} \label{tab.gen1}
\vspace{-1em}
\end{table*}

\subsection{Spatiotemporal IoU Loss}
Bounding box regression loss plays a key role in object detection. IoU, GIoU \cite{rezatofighi2019giou}, and CIoU \cite{zheng2021ciou} are widely used in frame-based object detection and we show the loss function of YOLOv4 in Eq. \ref{eq.yololoss}, where $\mathcal{L}_{obj}$ is confidence loss, $\mathcal{L}_{cls}$ is classification loss and $\mathcal{L}_{reg}$ is bounding box regression loss. Current event-based detection methods often directly apply these frame-based approaches, which compute the coordinate differences between the predicted boxes and gtboxes. This neglects the unique spatiotemporal characteristics of event-based spike trains, potentially leading to less precise regression.

\begin{equation}\label{eq.yololoss}
\begin{aligned}
\mathcal{L}_{total}&=\mathcal{L}_{obj}+\mathcal{L}_{cls}+\mathcal{L}_{reg} \\
\mathcal{L}_{reg}&=1-CIoU
\end{aligned}
\end{equation}

As previously noted, the events generated by the target objects exhibit spatiotemporal continuity. This can be represented through the spike density $\rho$ (total number of spikes within a box divided by its area). $\rho$ around the object is usually higher than around the background. We propose Spiking-IoU to quantify the difference in spike density between the predicted $B=(x,y,w,h)$ and ground truth $B^{gt}=(x^{gt},y^{gt},w^{gt},h^{gt})$ boxes (Eq. ~\ref{eq.spikIoU}).

\begin{equation}\label{eq.spikIoU}
\begin{aligned}
\text{Spiking-IoU}\!&=\!\left| \rho_{sd}^{gt}\!-\! \rho_{sd}\right|\!
= \!\left| \frac{\sum_{m=1}^{M}\delta_m^{gt}}{w^{gt}h^{gt}}\!-\!\frac{\sum_{n=1}^{N}\delta_n}{wh}      \right|
\end{aligned}
\end{equation}

Furthermore, we propose ST-IoU loss to fully exploit the spatiotemporal feature in the event-based detection (Eq. \ref{eq.stiou}). It incorporates the conventional IoU scheme to provide accurate coordinate information for fast convergence and Spiking-IoU to enhance detection precision with unique spatiotemporal information. $a$ and $b$ are constant weights.

\begin{equation}
\begin{aligned}\label{eq.stiou}
\text{ST-IoU}  &= a\times \text{Spiking-IoU} + b\times\text{CIoU} \\
\mathcal{L}_{reg} &=1-\text{ST-IoU}
\end{aligned}
\end{equation}

\section{Experiments}
We first describe the experimental settings. Then, the effective tests and ablation tests are conducted to verify the effectiveness and performance of the proposed methods. Additionally, we analyze \textit{CREST} in exploiting sparsity.

\subsection{Experimental Settings}
\subsubsection{Datasets.} We adopt the commonly used NCARs dataset for object recognition \cite{sironi2018hats}, Gen1 and PKU-Vidar-DVS dataset for object detection \cite{detournemire2020largescaleeventbaseddetection, li2022retinomorphic}. The details of these datasets are shown in \textbf{Supplementary Material D}.

\subsubsection{Implementation Details.}
Set $K=5$ and $\alpha=3$ for middle layers, with $\alpha=1$ for the input encoding layer in FSN-SNN models. All models are trained on a Tesla V100 GPU using the AdamW optimizer and StepLR scheduler. \underline{For recognition}, training spans 100 epochs with a batch size of 64, an initial learning rate of $1e-3$, and a weight decay of $1e-2$. \underline{For detection}, training also spans 100 epochs with a batch size of 32, an initial learning rate of $1e-3$, and a weight decay of $5e-4$. Data augmentation techniques include mosaic, random rotation, crop and flipping.

\subsubsection{Performance Metrics.} \textbf{Accuracy} \& \textbf{mAP$_{50:95}$ \& mAP$_{50}$} are used for object recognition and detection respectively. We also establish the \textbf{energy consumption model} to show the efficiency of \textit{CREST}. Present works only consider the energy consumed by spike integration, which neglect the energy required for the decay operation of LIF-like neurons despite its inescapability in hardware implementation
\cite{fan2024sfod,su2023deep,dampfhoffer2022snns}. 

\underline{For ANNs}, energy consumption is assessed by the number of MACs ($OP_{MAC}$) (Eq. \ref{eq.ANNenergy}). For fairness, we also account for sparsity ($sp$) in methods that leverage spatiotemporal sparsity \cite{messikommer2020event}. \underline{For LIF-based SNNs}, energy consumption includes spike integration and extra MACs caused by membrane potential decay (Eq. \ref{eq.LIF_SNNenergy}). \underline{For FSN-based SNNs}, as mentioned in Background and Motivation, they have negligible decay energy costs (Eq. \ref{eq.FS_SNNenergy}). The energy cost for per 32-bit floating-point AC/MAC operation is 0.9/4.6 pJ \cite{horowitz20141}.

\begin{align}
\label{eq.ANNenergy}E_{ANN} & =OP_{MAC} \times sp \times E_{MAC} \\
\label{eq.LIF_SNNenergy} E_{LIF-SNN} &= K \times OP_{AC} \times fr \times E_{AC} \notag\\ 
&\quad + K \times OP_{AC} \times (1 - fr) \times E_{MAC} \\
\label{eq.FS_SNNenergy}E_{FS-SNN}&=K\times OP_{AC} \times fr \times E_{AC} 
\end{align}

\subsection{Effective Tests}
\subsubsection{Object Recognition.} NCAR samples are resized to 64×64 pixels, encoded with MESTOR and conjointly-trained with three FSN-based networks \cite{bochkovskiy2020yolov4,huang2017densely,ma2018shufflenet}. Our model outperforms existing methods in accuracy with smaller networks in Table \ref{tab.ncar}. Moreover, its lower firing rate reduces energy consumption compared to sparse ANNs and SOTA SNNs.

\begin{table}[t]
\centering
\setlength\tabcolsep{1pt} 
\fontsize{9pt}{8pt}\selectfont
\begin{tabular}{ccccc}
\toprule
Representation &  Backbone &  K  & fr/sp & mAP$_{50}$\\
\midrule
HIST \cite{chen2018pseudo}  & YOLO &  ANN & 1 & 0.331\\

HIST \cite{iacono2018towards}  &  SSD &  ANN  &  1& 0.326\\
EV. \cite{hu2020learning}  &  YOLOv3 &  ANN & 1&  0.353\\
TAR-events \cite{li2022retinomorphic}  &  YOLOv3 &  ANN &  1& 0.386\\
\midrule
\multirow{3}{*}{MESTOR} & CSPdark & 5  & 0.173 &  \textbf{0.432}   \\
~ & Dense121 & 5  & \textbf{0.092}&  0.429  \\
~ & ShuffleV2 & 5 & 0.195&  0.425   \\
\bottomrule
\multicolumn{5}{l}{\makecell[l]{{\small{EV. denotes EventVolume.}}}}
\end{tabular}
\caption{Comparison between \textit{CREST} and the state-of-the-art methods on PKU-DVS-Vidar dataset.} \label{tab. PKU-DVS-Vidar}
\end{table}
\begin{table}[t]
\centering
\fontsize{9pt}{8pt}\selectfont
\setlength\tabcolsep{2.5pt} 
\begin{tabular}{ccccccc}
\toprule
Method &CLR &  MESTOR &  ST-IOU &  mAP$_{50:95}$ & fr &Time\\
\midrule
Baseline&  & & &  0.269   & 0.184 & $1\times$   \\ 
CREST-A &\CheckmarkBold&   &    &  0.279    & 0.185  & $3.42\times$ \\
CREST-B &\CheckmarkBold&\CheckmarkBold & &  0.358    & 0.173  &   $3.43\times$  \\ 
\textit{CREST} &\CheckmarkBold&\CheckmarkBold &\CheckmarkBold    & 0.360      & 0.167  & $3.41\times$ \\ 
\bottomrule
\end{tabular}
\caption{Ablation study: comparison between the baseline, CREST-A, CREST-B, and \textit{CREST} on Gen1.} \label{tab.ablationoverview}
\end{table}

\begin{table}[t]
\centering
\fontsize{9pt}{8pt}\selectfont
\begin{tabular}{ccccccccc}
\toprule
\multirow{2}{*}{Method} &  \multicolumn{3}{c}{Acc} \\\cmidrule(r){2-4} 
                     & SFOD & w/ FSN-BP(ours)& w/ CLR(ours)     \\ \midrule
DenseNet121-16  & 0.937& 0.948& 0.952   \\
DenseNet121-32  & 0.923& 0.948 & 0.953    \\
DenseNet169-16  & 0.921& 0.946& 0.950     \\
DenseNet169-32  & 0.894& 0.944 & 0.953   \\ 

\bottomrule
\end{tabular}
\caption{Ablation study: comparison between the conventional learning rule and the proposed CLR on NCARs.} \label{tab.ablationCLR}
\end{table}
\begin{table}[t]
\centering
\fontsize{9pt}{8pt}\selectfont
\setlength\tabcolsep{4pt} 
\begin{tabular}{ccccccc}
\toprule
\multirow{2}{*}{Representation} & \multicolumn{2}{c}{CSPdark-tiny} & \multicolumn{2}{c}{Dense121-16} & \multicolumn{2}{c}{ShuffleV2} \\ \cmidrule(r){2-3} \cmidrule(r){4-5} \cmidrule(r){6-7}
                        & Acc           & fr         & Acc           & fr        & Acc          & fr       \\
\midrule
MESTOR-A     & 0.873   & 0.164 & 0.871  & 0.151  & 0.859  & 0.187 \\
MESTOR-B    & 0.938  & 0.185   & 0.937   & 0.154  & 0.922  & 0.199 \\
MESTOR  & 0.949  & 0.165   & 0.952  & 0.146 & 0.940 & 0.176   \\
\bottomrule
\end{tabular}
\caption{Ablation study: effectiveness of MESTOR’s components on NCARs dataset.} \label{tab.ablationmestor}
\vspace{-0.5em}
\end{table}

\begin{figure}[tb]
  \centering 
  \includegraphics[width=\linewidth]{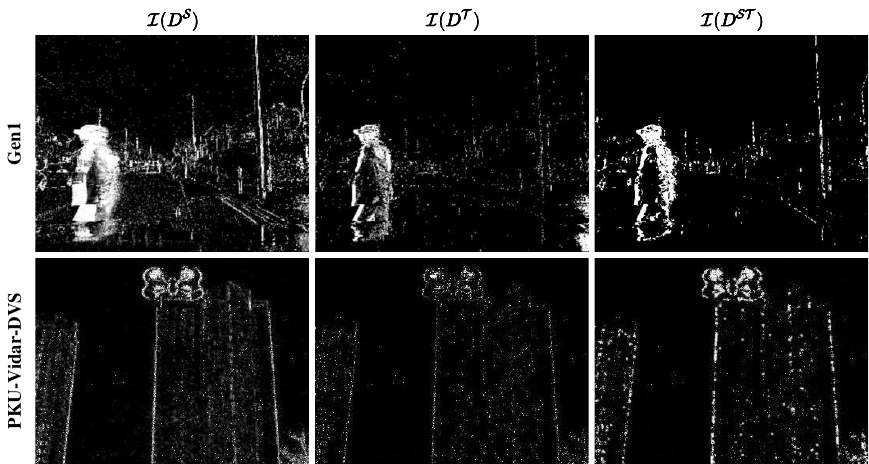} 
  \caption{Gen1/PKU-Vidar-DVS encoded with MESTOR.} 
  \label{fig.coding} 
\end{figure}

\begin{figure}[tb]
  \centering 
  \includegraphics[width=\linewidth]{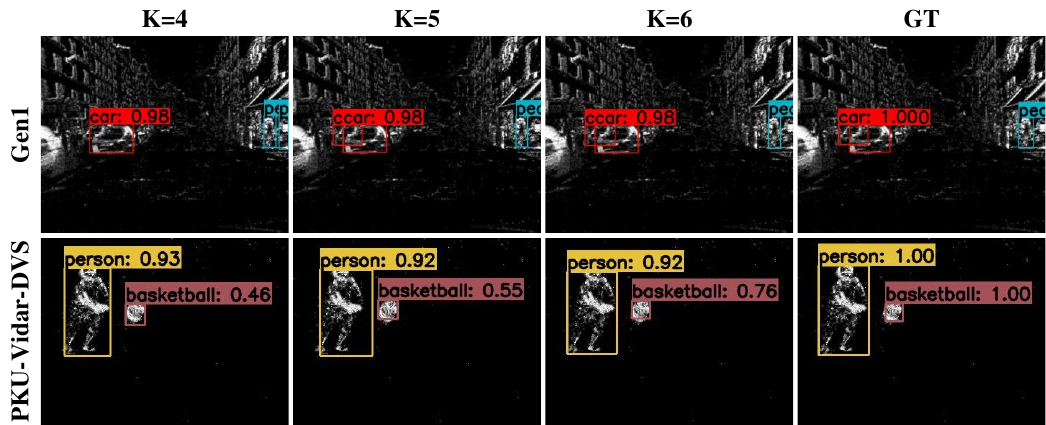} 
  \caption{Visualized detection results with different K.} 
  \label{fig.dec} 
\end{figure}

\subsubsection{Object Detection.}  We evaluate \textit{CREST} on the Gen1 and PKU-Vidar-DVS (9 classes under extreme lighting conditions for event-based object detection) datasets. The visualization of the MESTOR encoding process and the detection results are shown in Fig. \ref{fig.coding} and Fig. \ref{fig.dec}. As shown in Table \ref{tab.gen1}, \textit{CREST} achieves high performance with fewer parameters and ultra-low energy consumption. It improves $mAP_{50}$ from 0.59 to 0.63 compared to SOTA SNNs, while reducing energy consumption by $100\times$. Even our smallest network (1.79M parameters) performs competitively with existing SNNs, whereas the sparse ANN \cite{messikommer2020event} lag behind in mAP. Note that the highest mAP in \cite{zubic2023chaos} is achieved with a much larger network. In Table \ref{tab. PKU-DVS-Vidar}, \textit{CREST} performs well under these challenging conditions, surpassing existing ANN methods in mAP.

\subsection{Ablation Studies}
\subsubsection{Overview.} We conducted ablation studies to validate the effectiveness of each component in \textit{CREST} (Table \ref{tab.ablationoverview}). The baseline is trained with FSN-BP, $K\!\!=\!\!5$, CSPdarknet-tiny backbone, YOLOv4-head, CIOU regression loss, and histogram representation. CREST-A replaces FSN-BP in the baseline with CLR, and CREST-B further substitutes the representation in CREST-A with MESTOR.

The conjoint learning rule reduces training time per epoch by $3.4\times$ and enhances mAP performance by reducing errors caused by surrogate gradient backpropagation across multiple dimensions. MESTOR and ST-IoU further boost mAP and lower firing rates by fully leveraging the spatiotemporal feature of the event data.

\subsubsection{Conjoint Learning Rule.} We validate CLR's effectiveness in mitigating gradient vanishing and enhancing performance in deep SNNs (Table \ref{tab.ablationCLR}). LIF-based SFOD trained with STBP  \cite{wu2018stbp} struggle with deep SNNs, resulting in low accuracy. In contrast, our model trained with CLR demonstrates the accuracy improvement compared with both SFOD and our model trained with FSN-BP. 

\subsubsection{Multi-scale Spatiotemporal Event Integrator.} We denote spatiotemporal continuous $\mathcal{I}(D^{\mathcal{ST}})$ as A, and spatial $\mathcal{I}(D^{\mathcal{S}})$ \& temporal $\mathcal{I}(D^{\mathcal{T}})$ as B. Ablation experiments on MESTOR components (Table \ref{tab.ablationmestor}) show that A reduces the firing rate by reducing redundant events through extracting and clustering the spatiotemporal continuous events, while B improves accuracy by retaining key spatiotemporal features. Integrating A and B (MESTOR) achieves the highest accuracy and a comparatively low firing rate.

\subsubsection{Spatiotemporal-IoU.} 
\begin{figure}[t]
  \centering 
  \includegraphics[width=\linewidth]{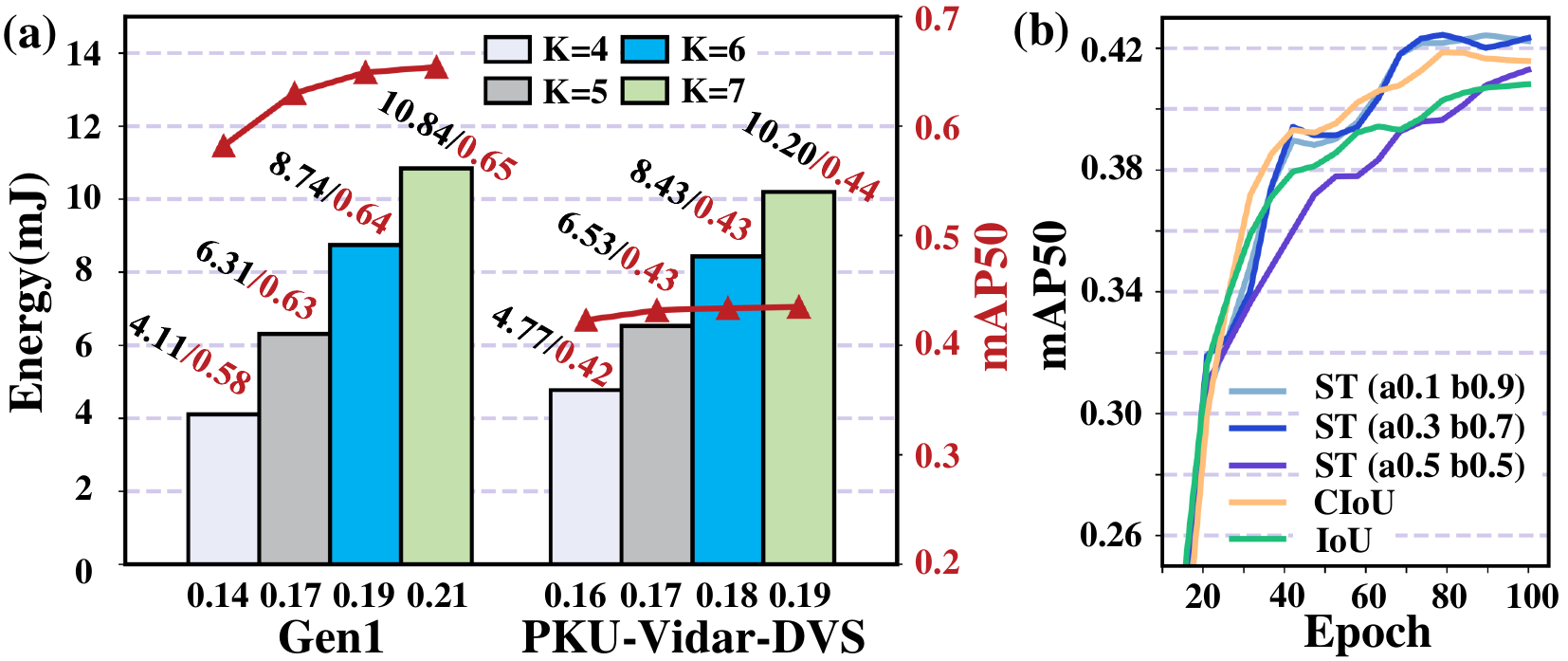} 
  \caption{ (a) Energy and mAP change with time-window on CSPdarknet-tiny ; (b) mAP of Shufflenet on PKU-Vidar-DVS with different IoUs.} 
  \label{fig.Tenergy} 
\end{figure}

Fig. \ref{fig.Tenergy}(b) shows \textit{CREST} with different IoUs. ST-IoU achieves higher mAP than CIoU. Experiments with different a\&b in Eq. \ref{eq.stiou} show that in ST-IoU, the CIoU part accelerates the convergence speed, while the Spiking-IoU part improves the mAP due to its ability to fully exploit the spatiotemporal feature of the event-based data.

\subsubsection{Additional analysis of \textit{CREST} in exploiting sparsity.} Fig. \ref{fig.Tenergy}(a) shows the energy consumption and accuracy of CSPdarknet-tiny on two datasets with varying degrees of sparsity (i.e. firing rates under different time windows). The energy cost is reduced by up to $53.2\%$ with nearly unpruned mAP when the firing rate decays.

\vspace{-0.25em}
\section{Conclusion}
We propose \textit{CREST}, an innovative conjointly-trained spike-driven framework (incorporate MESTOR, ST-IoU loss, and FSN-SNN model) tailored for high accuracy and energy-effcient event-based object detection. With up to 100$\times$ improvement in energy efficiency over SOTA SNN algorithms, \textit{CREST} offers a groundbreaking solution for advanced event-based object detection systems.

\section{Acknowledgments}
This work was supported by NSAF under Grant U2030204. (Corresponding author: Jun Zhou.)

\section{Supplementary Material}
\subsection{A. LIF Neuron and Its Variants} 
Leaky Integrate-and-Fire (LIF) neuron \cite{abbott1999lif} and its variants, including IF \cite{gerstner2002IF} and PLIF \cite{fang2021plif}, are most commonly used in SNN algorithms and hardware implementations, due to their trade-off between low computational complexity and biological interpretability.

These neuron mimic the membrane potential dynamics and the spiking scheme, encoding a real value into a spike train (ST) to implement energy-saving computing. The length of a spike train is called time window K.

\subsubsection{LIF neuron models.} For LIF neuron,  we consider the version of parameters dynamics that is discrete in time. LIF neuron $n$ in layer $l$ integrates all the input spikes $\delta_m^{l-1}(t)$ from layer $l\!-\!1$ and accumulates to the membrane potential $u_n^l(t)$ which also leaks at each timestep $t$ by a fixed factor $\tau$. A spike $\delta_n^l(t)$ is fired when the membrane potential exceeds threshold $U_{th}$, and the membrane potential will be reset to $U_{reset}$. Denote the weight between neuron $m$ and neuron $n$ as $w_{mn}$ and a discrete form of the LIF neuron can be characterized as follows:

\begin{align}
\label{eq.lif_integration}U_{n}^l(t) &=\sum_{m=1}^{N^{l-1}}w_{mn}\delta_m^{l-1}(t)\\
\label{eq.lif_spike}\delta_n^{l}(t) &= \begin{cases}1, &  u_n^l(t) \ge U_{th} \\ 0, & \text { otherwise }\end{cases} \\
\label{eq.lif_u_decay}u_{decay}^{l}(t) &= u_n^{l}(t)+(U_{n}^l(t) + U_{reset} - u_n^{l}(t))/ \tau \\
\label{eq.lif_u(t+1)}u_n^{l}(t+1) &= \begin{cases}u_{decay}^{l}(t), &\delta_n^{l}(t)=0 \\U_{reset}, & \text { otherwise }\end{cases} 
\end{align}

\subsubsection{PLIF neuron models.}
Parametric Leaky Integrate-and-Fire (PLIF) spiking neuron model have a similar function as LIF model. Futhermore, its membrane time constant $\tau$ is optimized automatically during training, rather than being set as a hyperparameter manually before training, increasing the heterogeneity of neurons. Specifically, PILF replaced $\tau$ with $1/k(a)$ in Eq. \ref{eq.lif_u_decay}. Moreover, $a$ is a trainable parameter.

\begin{align}
\label{eq.plif_k(a)}k(a) &= 1/(1+exp(-a))\\
\label{eq.plif_u_decay}u_{decay}^{l}(t) &= u_n^{l}(t)+k(a)(U_{n}^l(t) + U_{reset} - u_n^{l}(t)) 
\end{align}

\subsubsection{IF neuron models.}
IF model not only remove the potential leaky item in Eq. \ref{eq.lif_u_decay} , but also set $U_{reset}=0$ to simplify the computation process. 

\begin{align}
\label{eq.if_u_decay}u_{decay}^{l}(t) &= u_n^{l}(t) \\
\label{eq.if_u(t+1)}u_n^{l}(t+1) &= \begin{cases}u_{decay}^{l}(t), &\delta_n^{l}(t)=0 \\0, & \text { otherwise }\end{cases} 
\end{align}

\subsection{B. Choice between ReLU and Leaky-ReLU}
ANN-based object detection mostly adopts ReLU and leaky-ReLU \cite{glorot2011relu, maas2013leakrelu}. Leaky-ReLU allows a small, positive gradient $\beta_{neg}$ when the input is negative, and the gradient $\beta$ is 1 when the input is positive. To emulate it with FSN (FSN-LReLU), we revise the FSN spiking function as follows:

\begin{align}
\label{eq.integration_sup}U_{n}^l &=\sum_{m=1}^{N^{l-1}}\sum_{t=1}^{K^{l-1}}d^{l-1}(t)\beta^{l-1}\delta_m^{l-1}(t)w_{mn}^l\\
\label{eq.spikeileak_sup}\delta_n^l(t) &= \begin{cases}1, & u_n^l(t) \ge U_{th}^{l}(t) \\-1, &  u_n^l(t) \le -U_{th}^{l}(t) \\ 0, & \text {otherwise }\end{cases}\\
\label{eq.leakb_sup}\beta^l &= \begin{cases}\beta_{neg}, & \delta_n^l(t) < 0 \\1,   &\text {otherwise }\end{cases}\\
\label{eq.membraneDecay_sup}u_n^l(t+1)&=U_n^l-\sum_tU_{th}^{l}(t)\delta_n^l(t)   
\end{align}

This time, FSN emulates leaky-ReLU without error when receiving inputs from $\{-\alpha(1-2^{-K}), \dots, -\alpha2^{-K}\}\cup\{\alpha2^{-K}, \alpha2^{1-K}, \dots, \alpha(1-2^{-K})\}$ and rounds down values between adjacent levels. \cite{mao2024stellar} introduced an effective FSN-ReLU encoder that extracts the spike trains from the integrated membrane potential with AND gates and multiplexers, reducing computational complexity by eliminating iterative subtractions ($K$ times for each neuron) in Eq. \ref{eq.membraneDecay_sup}. In contrast, FSN-LReLU needs extra encoding and decoding hardware modules to convert between membrane potentials and signed spike trains, increasing hardware costs. 

Furthermore, with the same experimental setting as the main paper, Table \ref{tab.leakyrelu} indicates that SNN based on FSN-ReLU and FSN-LReLU achieve similar accuracy. But FSN-LReLU generates extra spikes when the membrane potential is negative, which leads to higher energy consumption.
\begin{table}[htb]
\centering
\setlength\tabcolsep{2pt} 
\begin{tabular}{ccccccc}
\toprule
\multirow{2}{*}{Neuron} & \multicolumn{2}{c}{CSPdark-tiny} & \multicolumn{2}{c}{Dense121-16} & \multicolumn{2}{c}{ShuffleV2} \\ \cmidrule(r){2-3} \cmidrule(r){4-5} \cmidrule(r){6-7}
                        & Acc           & fr         & Acc           & fr        & Acc          & fr       \\
\midrule
FSN-ReLU     & 0.949   & 0.165 & 0.952  & 0.146  & 0.940  & 0.176 \\
FSN-LReLu    & 0.949  & 0.310   & 0.952  & 0.181  & 0.942   & 0.290 \\
\bottomrule
\end{tabular}
\caption{Comparison between the FSN-ReLU and FSN-LReLu ($\beta_{neg}=0.1$) on NCARs dataset.} \label{tab.leakyrelu}
\end{table}

This suggests that sufficient robustness brought by spike encoding makes the choice between Leaky ReLU and ReLU less significant in terms of  accuracy. In this way, we choose FSN-ReLU as our fundamental neuron for energy efficiency.

\subsection{C. Backpropagation Principles in DL-Net}
As mentioned in the main paper, the working pattern of DL-Net is described by the following equation:

{\begin{equation}\label{eq.dln_clip_sup}
x_q^{l-1}=\text{clip}(x^{l-1}_m)=\begin{cases} 
0, &  x^{l-1}_m < 0 \\
x^{l-1}_m, &  0\leq x^{l-1}_m\leq X^{l-1} \\
X^{l-1}, & \text{otherwise}
\end{cases}
\end{equation}}

{\begin{equation}\label{eq.dln_quantilize_sup}
x_p^{l-1}= \text{round}(\frac{x_q^{l-1}}{X^{l-1}_\text{min}})\times X^{l-1}_\text{min}
\end{equation}}

{\begin{equation}\label{eq.dln_integrate_sup}
x^l_n=\sum_{p=1}^{N^{l-1}}x_p^{l-1}w_{pn}^l
\end{equation}}

Taking the object recognition task as an example, the backpropagation principles in DL-Net are illustrated as below. For the last layer $L$, the index of the neuron with the largest membrane potential corresponds to the predicted output label. The loss function $\mathcal{L}$ is defined as:

\begin{align}
\label{eq.lossfunc} 
 \mathcal{L}=-\sum_{n=1}^{N^{L}}y_{n}log\sigma(x_{n}^{L})
\end{align}

where $y_{n}$ is the one-hot target label, $N^{L}$ is the number of neurons in layer L, $x_{n}^{L}$ denotes the membrane potential of neuron $n$ in $L$ and $\sigma(.)$ is the Softmax function. 

The gradient of  $\mathcal{L}$ to $x_{n}^{L}$ can be represented as:

\begin{align}
\label{eq.gradientL} 
\epsilon_{n}^{L}=\frac{\partial \mathcal{L}}{\partial x_{n}^{L}}=
\frac{\partial\mathcal{L}}{\partial \sigma(x_{n}^{L})}\frac{\partial\sigma(x_{n}^{L})}{\partial x_{n}^{L}}
=\sigma(x_{n}^{L})-y_{n}
\end{align}

Then, from Eq. \ref{eq.dln_clip_sup},\ref{eq.dln_quantilize_sup},\ref{eq.dln_integrate_sup}, the gradient of $w_{pn}^{L}$ can be obtained as:

\begin{align}
\label{eq.gradientLw} 
\frac{\partial \mathcal{L}}{\partial w_{pn}^{L}}= 
\frac{\partial \mathcal{L}}{\partial x_{n}^{L}}
\frac{\partial x_{n}^{L}}{\partial w_{pn}^{L}}
=\epsilon_{n}^{L}\frac{\partial x_{p}^{L-1}w_{pn}^{L}}{\partial w_{pn}^{L}}
=\epsilon_{n}^{L}x_{p}^{L-1}
\end{align}

Note that $\frac{\partial x_{p}^{l-1}}{\partial x_{q}^{l-1}}$ is estimated to be 1 in DL-Net, the error passed from neuron $n$ in layer $L$ to neuron $m$ in layer $L-1$ is denoted as:

\begin{align}
\label{eq.gradientL2L-1} 
\epsilon_{m}^{L-1} = \frac{\partial  \mathcal{L}}{\partial x_{m}^{L-1}} & =
\sum_{n=1}^{N^{L}} \frac{\partial x_{n}^{L}}{\partial x_{p}^{L-1}}
\frac{\partial x_{p}^{L-1}}{\partial x_{q}^{L-1}}
\frac{\partial x_{q}^{L-1}}{\partial x_{m}^{L-1}}\epsilon_{n}^{L} \notag\\ 
&\quad= \sum_{n=1}^{N^{L}} w_{pn}^{L} \frac{\partial x_{q}^{L-1}}{\partial x_{m}^{L-1}}\epsilon_{n}^{L}
\end{align}

And from Eq. \ref{eq.dln_clip_sup}, the gradient of $x_{q}^{L-1}$ to $x_{m}^{L-1}$ can be obtained as:

\begin{align}
\label{eq.gradientq2m} 
\frac{\partial x_{q}^{L-1}}{\partial x_{m}^{L-1}}=rect(\frac{2x_{m}^{L-1}-X^{L}}{2X^{L}} )
\end{align}

Then, the gradient of $w_{gm}^{L-1}$ can be denoted as:

\begin{align}
\label{eq.gradientL2L-2} 
\frac{\partial \mathcal{L}}{\partial w_{gm}^{L-1}}= 
\frac{\partial \mathcal{L}}{\partial x_{m}^{L-1}}
\frac{\partial x_{m}^{L-1}}{\partial w_{gm}^{L-1}}=\epsilon_{m}^{L-1}x_{p}^{L-2}
\end{align}

In this way, the gradient of weights in other layers can be calculated similarly.

\subsection{D. Datasets details}

\subsubsection{NCARs dataset.}
The NCARs dataset \cite{sironi2018hats} are choosed for event-based object recognition task, which comprising 12,336 car samples and 11,693 background samples. Each sample has a duration of 100 ms and exhibits varying spatial dimensions.
\subsubsection{Gen1 dataset.}
The Gen1 dataset \cite{detournemire2020largescaleeventbaseddetection} is a large-scale  dataset widely used for event-based object detection. It consists of 39 hours of recordings with more than 228k car and 28k pedestrian annotations in driving scenarios. Bounding box labels for cars and pedestrians within the recordings are provided at frequencies between 1 to 4Hz.
\subsubsection{PKU-Vidar-DVS dataset.}
PKU-Vidar-DVS dataset \cite{li2022retinomorphic} is another event-based object detection dataset, involving high-speed and low-light scenes contains 9 indoor and outdoor challenging scenarios. Annotations in the recordings are provided at a frequency of 50 Hz. dataset has 99.6k labeled timestamps and 215.5k labels in total. Only DVS part of the PKU dataset are  used in this paper.
\bibliography{aaai25}
\end{document}